\documentclass[report,camera-ready]{deepleap-paper}

\title{DELE-w0.5: Inferring Action from Future Latent State for Robotic Manipulation}
\subtitle{}
\shorttitle{One Intelligence Across Embodiments}
\papertype{Technical Report}
\author{Fenghao Lei \quad Zhixiong Huang \quad Long Yang \quad Jiabao Chen 
 Peilin Huang\\
  \quad Cong Fang \quad Han Fu \quad Zhuo Li \quad Xiaoxue Ren}
\affiliations{DeepLeap Technology Co., Ltd., Shenzhen, China\\
 Email:~yanglong0908@gmail.com\\
 Official~Page: \url{https://deepleap-x.com/research/dele-w0.5}
}
\paperabstract{%
 World-Action Models (WAMs) build robot control on video-generation backbones, which jointly predict dense future visual trajectories and robot actions.
We argue that video generation is an unnecessary intermediate objective for world-action modeling.
For robotic manipulation, the goal of a world model is not to reproduce how the world looks at every intermediate moment, but to predict the state that the world will reach after an action is executed.
The intermediate frames only describe the visual transition between physical states, which consumes substantial model capacity and computation, but do not directly specify the physical outcome that the robot action is intended to produce.
In this paper, we propose $\mathtt{DELE}$-$\mathtt{w0.5}$, which infers robot actions from predicted future states without relying on video generation.
Concretely, $\mathtt{DELE}$-$\mathtt{w0.5}$  infers the action sequence from its corresponding compact future latent state.
The future latent state captures the action-relevant physical outcome of robot interaction and serves as an explicit bridge between world modeling and action generation.
The core design principle of $\mathtt{DELE}$-$\mathtt{w0.5}$ is to model how the physical world changes under robot actions, rather than how its visual appearance evolves frame by frame.
This formulation removes the high-dimensional visual redundancy introduced by dense video representations, and it therefore enables cheaper training and low-latency inference.
Across 640 real-robot trials on four long-horizon manipulation tasks, our $\mathtt{DELE}$-$\mathtt{w0.5}$ achieves the best performance among all compared policies, attaining 62.5\% overall full-task success and 81.3\% macro ordered-stage progress. It outperforms the strongest baseline by 32.5 percentage points in full-task success and 20.1 percentage points in macro progress.
}
\codeurl{https://deepleap-x.com/research/dele}
\pdfauthorinfo{DeepLeap Technology Co., Ltd.}

\begin{document}
\maketitle
\deepleapcontents

\section{Introduction}

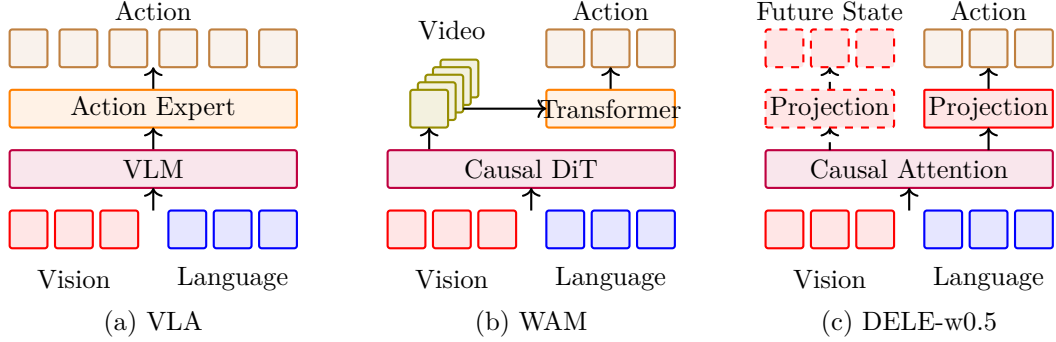
\begin{figure}[t]
\centering

\begin{tikzpicture}
[line width=0.8pt,scale=1.0] 

\node at (1.9,0.5+2.4+0.25) {Action};


\draw[
    rounded corners=1pt,    
    fill=brown!10,           
    draw=brown  
] (0,2.4) rectangle (0.5,0.5+2.4);

\draw[
    rounded corners=1pt,    
    fill=brown!10,           
    draw=brown   
] (0.5+0.16,2.4) rectangle (0.5+0.16+0.5,0.5+2.4);

\draw[
    rounded corners=1pt,    
    fill=brown!10,           
    draw=brown    
] (0.5+0.16*2+0.5,2.4) rectangle (0.5+0.16*2+0.5*2,0.5+2.4);

\draw[
    rounded corners=1pt,    
    fill=brown!10,           
    draw=brown    
] (0.5+0.16*3+0.5*2,2.4) rectangle (0.5+0.16*3+0.5*3,0.5+2.4);

\draw[
    rounded corners=1pt,    
    fill=brown!10,           
    draw=brown    
] (0.5+0.16*4+0.5*3,2.4) rectangle (0.5+0.16*4+0.5*4,0.5+2.4);

\draw[
    rounded corners=1pt,    
    fill=brown!10,           
    draw=brown   
] (0.3+0.6*5,2.4) rectangle (0.8+0.6*5,0.5+2.4);

\draw[->,black] (1.9,2.1)--(1.9,2.4);

\draw[
    rounded corners=1pt,    
    fill=orange!10,           
    draw=orange  
] (0,1.6) rectangle (0.8+0.6*5,2.1);

\node at (1.9,1.85) {Action Expert};

\draw[->,black] (1.9,1.3)--(1.9,1.6);

\draw[
    rounded corners=1pt,    
    fill=purple!10,           
    draw=purple   
] (0,0.8) rectangle (0.8+0.6*5,1.3);

\node at (1.9,1.05) {VLM};

\draw[->,black] (1.9,0.5)--(1.9,0.8);

\draw[
    rounded corners=1pt,    
    fill=red!10,           
    draw=red    
] (0,0) rectangle (0.5,0.5);

\draw[
    rounded corners=1pt,    
    fill=red!10,           
    draw=red   
] (0.6,0) rectangle (0.5+0.6,0.5);

\draw[
    rounded corners=1pt,    
    fill=red!10,           
    draw=red    
] (0.6*2,0) rectangle (0.5+0.6*2,0.5);

\node at (2.95,-0.4) {Language};


\draw[
    rounded corners=1pt,    
    fill=blue!10,           
    draw=blue  
] (0.3+0.6*3,0) rectangle (0.8+0.6*3,0.5);

\draw[
    rounded corners=1pt,    
    fill=blue!10,           
    draw=blue   
] (0.3+0.6*4,0) rectangle (0.8+0.6*4,0.5);

\draw[
    rounded corners=1pt,    
    fill=blue!10,           
    draw=blue   
] (0.3+0.6*5,0) rectangle (0.8+0.6*5,0.5);

\node at (0.85,-0.4) {Vision};

\node at (1.9,-1) {(a)~VLA};


\node at (0.85+5,0.5+2.4) {Video};

\node at (2.95+5,0.5+2.4+0.25) {Action};

\draw[
    rounded corners=1pt,    
    fill=brown!10,           
    draw=brown  
] (0.3+0.6*3+5,0+2.4) rectangle (0.8+0.6*3+5,0.5+2.4);

\draw[
    rounded corners=1pt,    
    fill=brown!10,           
    draw=brown   
] (0.3+0.6*4+5,0+2.4) rectangle (0.8+0.6*4+5,0.5+2.4);

\draw[
    rounded corners=1pt,    
    fill=brown!10,           
    draw=brown  
] (0.3+0.6*5+5,0+2.4) rectangle (0.8+0.6*5+5,0.5+2.4);

\draw[->,black] (0.6*4+5+0.55,2.1)--(0.6*4+5+0.55,2.4);


\draw[
    rounded corners=1pt,    
    fill=olive!10,           
    draw=olive    
] (0+0.1*4+5.3,0+1.6+0.1*4) rectangle (0.5+0.1*4+5.3,2.1+0.1*4);

\draw[
    rounded corners=1pt,    
    fill=olive!10,           
    draw=olive    
] (0+0.1*3+5.3,0+1.6+0.1*3) rectangle (0.5+0.1*3+5.3,2.1+0.1*3);

\draw[
    rounded corners=1pt,    
    fill=olive!10,           
    draw=olive    
] (0+0.1*2+5.3,0+1.6+0.1*2) rectangle (0.5+0.1*2+5.3,2.1+0.1*2);

\draw[
    rounded corners=1pt,    
    fill=olive!10,           
    draw=olive   
] (0+0.1+5.3,0+1.6+0.1) rectangle (0.5+0.1+5.3,2.1+0.1);

\draw[
    rounded corners=1pt,    
    fill=olive!10,           
    draw=olive    
] (0+5.3,0+1.6) rectangle (0.5+5.3,2.1);

\draw[
    rounded corners=1pt,    
    fill=orange!10,           
    draw=orange  
] (0.3+0.6*3+5,1.6) rectangle (0.8+0.6*5+5,2.1);

\node at (7.95,1.85) {Transformer};

\draw[->,black] (1.55+4,1.3)--(1.55+4,1.6);

\draw[->,black] (6,1.6+0.25)--(0.3+0.6*3+5,1.6+0.25);

\draw[
    rounded corners=1pt,    
    fill=purple!10,           
    draw=purple   
] (0+5,0.8) rectangle (0.8+0.6*5+5,1.3);

\node at (1.9+5,1.05) {Causal~DiT};

\draw[->,black] (1.9+5,0.5)--(1.9+5,0.8);

\draw[
    rounded corners=1pt,    
    fill=red!10,           
    draw=red    
] (0+5,0) rectangle (0.5+5,0.5);

\draw[
    rounded corners=1pt,    
    fill=red!10,           
    draw=red   
] (0.6+5,0) rectangle (0.5+0.6+5,0.5);

\draw[
    rounded corners=1pt,    
    fill=red!10,           
    draw=red    
] (0.6*2+5,0) rectangle (0.5+0.6*2+5,0.5);

\node at (0.85+5,-0.4) {Vision};


\draw[
    rounded corners=1pt,    
    fill=blue!10,           
    draw=blue  
] (0.3+0.6*3+5,0) rectangle (0.8+0.6*3+5,0.5);

\draw[
    rounded corners=1pt,    
    fill=blue!10,           
    draw=blue   
] (0.3+0.6*4+5,0) rectangle (0.8+0.6*4+5,0.5);

\draw[
    rounded corners=1pt,    
    fill=blue!10,           
    draw=blue   
] (0.3+0.6*5+5,0) rectangle (0.8+0.6*5+5,0.5);

\node at (2.95+5,-0.4) {Language};

\node at (1.9+5,-1) {(b)~WAM};


\node at (0.85+5*2,0.5+2.4+0.25) {Future~State};

\node at (2.95+5*2,0.5+2.4+0.25) {Action};

\draw[dashed,
    rounded corners=1pt,    
    fill=red!10,           
    draw=red    
] (0+5*2,0+2.4) rectangle (0.5+5*2,0.5+2.4);

\draw[dashed,
    rounded corners=1pt,    
    fill=red!10,           
    draw=red   
] (0.6+5*2,0+2.4) rectangle (0.5+0.6+5*2,0.5+2.4);

\draw[dashed,
    rounded corners=1pt,    
    fill=red!10,           
    draw=red    
] (0.6*2+5*2,0+2.4) rectangle (0.5+0.6*2+5*2,0.5+2.4);

\draw[->,black] (2.95+5*2,2.1)--(2.95+5*2,2.4);

\draw[
    rounded corners=1pt,    
    fill=brown!10,           
    draw=brown  
] (0.3+0.6*3+5*2,0+2.4) rectangle (0.8+0.6*3+5*2,0.5+2.4);

\draw[
    rounded corners=1pt,    
    fill=brown!10,           
    draw=brown   
] (0.3+0.6*4+5*2,0+2.4) rectangle (0.8+0.6*4+5*2,0.5+2.4);

\draw[
    rounded corners=1pt,    
    fill=brown!10,           
    draw=brown  
] (0.3+0.6*5+5*2,0+2.4) rectangle (0.8+0.6*5+5*2,0.5+2.4);

\draw[->,dashed,black] (0.85+5*2,2.1)--(0.85+5*2,2.4);

\draw[ dashed,
    rounded corners=1pt,    
    fill=red!10,           
    draw=red  
] (0+5*2,1.6) rectangle (0.5+0.6*2+5*2,2.1);
\node at (10.85,1.85) {Projection};

\draw[
    rounded corners=0.5pt,    
    fill=red!10,           
    draw=red  
] (0.3+0.6*3+5*2,1.6) rectangle (0.8+0.6*5+5*2,2.1);
\node at (12.95,1.85) {Projection};

\draw[
    rounded corners=1pt,    
    fill=purple!10,           
    draw=purple   
] (0+5*2,0.8) rectangle (0.8+0.6*5+5*2,1.3);

\node at (1.9+5*2,1.05) {Causal Attention};

\draw[->,dashed,black] (0.85+5*2,1.3)--(0.85+5*2,1.6);
\draw[->,black] (2.95+5*2,1.3)--(2.95+5*2,1.6);

\draw[->,black] (1.9+5*2,0.5)--(1.9+5*2,0.8);

\draw[
    rounded corners=1pt,    
    fill=red!10,           
    draw=red    
] (0+5*2,0) rectangle (0.5+5*2,0.5);

\draw[
    rounded corners=1pt,    
    fill=red!10,           
    draw=red   
] (0.6+5*2,0) rectangle (0.5+0.6+5*2,0.5);

\draw[
    rounded corners=1pt,    
    fill=red!10,           
    draw=red    
] (0.6*2+5*2,0) rectangle (0.5+0.6*2+5*2,0.5);

\node at (0.85+5*2,-0.4) {Vision};


\draw[
    rounded corners=1pt,    
    fill=blue!10,           
    draw=blue  
] (0.3+0.6*3+5*2,0) rectangle (0.8+0.6*3+5*2,0.5);

\draw[
    rounded corners=1pt,    
    fill=blue!10,           
    draw=blue   
] (0.3+0.6*4+5*2,0) rectangle (0.8+0.6*4+5*2,0.5);

\draw[
    rounded corners=1pt,    
    fill=blue!10,           
    draw=blue   
] (0.3+0.6*5+5*2,0) rectangle (0.8+0.6*5+5*2,0.5);

\node at (2.95+5*2,-0.4) {Language};

\node at (1.9+5*2,-1) {(c)~DELE-w0.5};

\end{tikzpicture}

\caption{Different Models for Manipulation.}
\label{fig:framework-comparsion}
\end{figure}

Embodied intelligence has achieved rapid development and achieved strong performance for robotic manipulation \cite{li2025developments,li2026rynnbrain}.
Vision-Language-Action (VLA) and World Action Model (WAM) are two popular model architectures.
VLA is a three-step framework, see Figure \ref{fig:framework-comparsion} (a): First, the input comprises multimodal data, including human task instructions and the robot's current observations;
Then, a vision-language model (VLM) backbone performs the task planning according to the input data;
Finally, an action expert decodes the task planning to the commands that are executable by the robot.
For some classic work of VLA, refer to \cite{brohan2022rt,pmlr-v229-zitkovich23a,kim2024openvla,intelligence2025pi05}.
Since the information density of language is lower than that of vision, VLA models require a large amount of data to adapt to the downstream tasks. 
As a result, generalization of VLA is poor, when light, color, or the task changes, the success rate decays rapidly, extensive empirical results have shown this view, see \cite{ye2026dreamzero,kim2026cosmospolicy}.
WAM \cite{UniPi2023,ye2026dreamzero,kim2026cosmospolicy} fundamentally discards the framework of VLA, WAM's core lying in the construction of a unified end-to-end learning objective, which integrates the prediction of world state evolution and action generation, see Figure \ref{fig:framework-comparsion} (b). The critical technology of WAM relies on self-supervised future state prediction on large-scale video data, 
then WAM develops the action generation as a conditional denoising process according to the future state prediction.

\subsection{Limitation of WAM}

A dominant line of recent World-Action Models (WAMs) builds robot control on top of video generation and jointly predicts future visual trajectories and actions \cite{ye2026dreamzero, yuan2026fastwam},
which brings some useful spatiotemporal priors into the policy. However, it also creates a structural mismatch between the objective of visual generation and the objective of robot control. Future videos are high-dimensional and appearance-sensitive. Most intermediate frames describe how the scene looks during a transition, rather than the physical consequence that determines the subsequent action. Therefore, the model is required to solve a substantially harder problem than control itself. It must first reconstruct dense visual evolution and then extract a compact action signal from it. Additionally, multi-frame video latents are much larger than the corresponding robot action sequence \cite{zhang2026imagewam}. 
They dominate the sequence length, memory consumption, and computational budget of joint world-action learning. Consequently, WAM training inherits the heavy cost of large video generators. Recent methods introduce compact backbones, latent downsampling, or lightweight adaptation to make video-action co-training more practical \cite{li2026lightwam}. These techniques reduce the cost of the existing pipeline, but do not remove its underlying redundancy. 
The same limitation becomes more critical during inference. Before an action can be executed, the model must perform iterative denoising over high-dimensional future video-action latents \cite{akbari2026flashwam}. The control loop is therefore bottlenecked by synthesizing a visual trajectory that is not itself executed by the robot \cite{li2026efficientwam}.

Therefore, the fundamental limitation of video-generation-based WAMs is not merely their computational cost. The deeper problem is that they optimize visual trajectory reconstruction as a surrogate for action-relevant world prediction. For robot control, a world model should predict how the world will become after an interaction, rather than reproduce how the world looks at every intermediate moment. This motivates a more direct formulation that models action-relevant future states and infers robot actions without relying on video generation in the control loop.

\subsection{Our Work}

We reformulate the current WAM of visual world modeling for robotic manipulation. 
It is different from recent video generation techinique based WAMs that model the future world through synthesizing dense visual trajectories, the proposed $\mathtt{DELE}$-$\mathtt{w0.5}$ argues that robot control does not require reconstructing the entire visual evolution of the real world, but only requires to predict the future latent state with downstream actions. With this observation, we formulate WAM as prediction problem of a future latent state and action, where the $\mathtt{DELE}$-$\mathtt{w0.5}$ directly infers the future visual state and corresponding robot behaviors, without any generating intermediate video frames.

\textbf{Rethinking World Model for Robotic Manipulation}.
We argue that existing video-generation-based WAMs formulate unnecessary intermediate objectives for robotic manipulation.
Although predicting future videos provides rich spatiotemporal priors, it implicitly treats visual evolution as the primary target of world modeling.
However, for robot robotic manipulation, a world model should not reproduce how the world looks at every intermediate moment.
Instead, it should predict how the world will become after an action is executed.
Most intermediate frames in generated videos only describe the visual transition between states, which provides limited information about the final physical consequence that determines subsequent actions.
Therefore, video-generation-based WAMs force the model to learn the distribution of visual evolution rather than directly modeling action-relevant future states.
This design introduces substantial computational overhead without necessarily improving control capability.
We argue that future-state prediction with causal relevance to robot actions is a more fundamental objective for embodied world models than reconstructing complete visual trajectories.

\textbf{Inferring Action from Future-State}.
With this observation, we propose $\mathtt{DELE}$-$\mathtt{w0.5}$ that infers robot actions from predicted future states rather than generating future visual trajectories.
Different from existing video-generation-based approaches that model the continuous evolution of the visual world, $\mathtt{DELE}$-$\mathtt{w0.5}$ treats future prediction as a state transition problem and focuses on the physical states that are causally relevant to subsequent actions.
The key insight is that a world model for robotic manipulation does not need to reproduce how the world visually evolves at every intermediate moment, but should capture how the world will become after an action is executed.
By directly modeling action-relevant future states, $\mathtt{DELE}$-$\mathtt{w0.5}$ avoids the high-dimensional visual redundancy introduced by video representations, and eliminates the costly iterative generation process required by video-based WAMs.
This formulation provides a more compact and efficient paradigm for world-action modeling, enabling significantly cheaper training and low-latency inference for real-world robotic deployment. Finally, we comparse the three formulation from probabilistic inference as follows,
\begin{itemize}
\item  VLA infers $p_{\theta}(\bA_{t}|\bO_{t},\bq_{t},\bL)$;
\item  WAM infers  $p_{\theta}(\bA_{t},\bO_{t:t+\Delta t}|\bO_{t},\bq_{t},\bL)$;
\item $\mathtt{DELE}$-$\mathtt{w0.5}$  infers $p_{\theta}(\bA_{t},\bO_{t+\Delta t}|\bO_{t},\bq_{t},\bL)$,
\end{itemize}
where $\bO_{t:t+\Delta t}$ is the video from current observation $\bO_{t}$ to the predicted future state $\bO_{t+\Delta t}$.

\section{Related Work}
\label{sec:related_work}

\subsection{Vision-Language-Action Models}

VLA models connect the semantic knowledge of pretrained vision-language models with low-level robot control. PaLM-E \cite{driess2023palme} injects continuous visual and state observations into a large language model for embodied reasoning. RT-2 \cite{pmlr-v229-zitkovich23a} further casts robot actions as language-like tokens and co-trains web-scale vision-language tasks with robot trajectories. Open X-Embodiment and RT-X \cite{oneill2023openx} show that heterogeneous trajectories from many robot platforms can support positive cross-embodiment transfer . 
Octo \cite{octo2024} provides an open generalist policy that supports different observations, action spaces, and robot embodiments. OpenVLA \cite{kim2024openvla} adapts a pretrained VLM to autoregressive action prediction and makes large-scale VLA training accessible. TinyVLA reduces model size and data requirements, while SmolVLA further targets training and deployment on affordable hardware \cite{wen2025tinyvla,shukor2025smolvla}.
RDT-1B \cite{liu2025rdt1b} uses a diffusion transformer for multi-modal bimanual actions. CogACT \cite{li2024cogact} couples a VLM with a diffusion action module. $\pi_0$  \cite{black2024pi0} instead uses flow matching to generate continuous action chunks. DexVLA \cite{wen2025dexvla} adopt dual-system designs in which a vision-language component provides semantic features and a diffusion expert produces motor commands.
FAST  \cite{pertsch2025fast} compresses high-frequency action sequences into a shorter autoregressive vocabulary. OpenVLA-OFT \cite{kim2025openvlaoft} uses parallel action decoding, continuous actions, and action chunking to improve both control rate and downstream success. UniVLA \cite{wang2025univla} learns task-centric latent actions from heterogeneous videos and decodes them for different embodiments. $\pi_{0.5}$ \cite{intelligence2025pi05} combines robot data, web data, semantic prediction, and heterogeneous supervision for open-world manipulation. HAMLET introduces compact historical memory, while ProgressVLA explicitly estimates task progress for long-horizon control \cite{koo2026hamlet,yan2026progressvla}. Qwen-VLA further unifies manipulation, navigation, and trajectory prediction across tasks and robot embodiments \cite{wang2026qwenvla}.

VLA still learns a direct conditional mapping from the current context to an action sequence, which is effective for behavior cloning, but it does not require the policy to represent the physical consequence of its action. 
The cost of large VLM backbones and iterative generative action heads also complicates real-time deployment. In contrast, $\mathtt{DELE}$-$\mathtt{w0.5}$ proposes a predicted future state as an explicit interface between perception and control. It first estimates the action-relevant physical outcome and then infers the robot action from this state. 

\subsection{World-Action Models}

World models learn predictive structure that can support control, planning, or policy learning, and recent studies show that future prediction can provide useful structure beyond a purely reactive policy.  Latent world models such as DreamerV3 \cite{hafner2025mastering} improve a policy through imagined state trajectories. 
Multi-view Masked World Models \cite{pmlr-v202-seo23a} learn predictive representations from multiple cameras for visual manipulation. 
UniPi \cite{UniPi2023} formulates sequential decision making as text-conditioned video generation and recovers actions from generated plans. 
Genie \cite{bruce2024genie} learns action-controllable interactive environments from unlabeled videos through latent actions. 
Recent robotic world models connect future prediction more tightly with action generation. GR-1 \cite{wu2023gr1} jointly predicts future images and robot actions after large-scale video pretraining. GR-2 \cite{cheang2024gr2} scales this paradigm with web videos and robot trajectories. 
3D-VLA \cite{zhen2024-3dvla} predicts goal images and point clouds to connect 3D reasoning with action planning. 
IRASim generates action-conditioned videos with frame-level action alignment for fine-grained robot interactions \cite{zhu2025IRASim}. 
Seer closes the loop through a predictive inverse-dynamics formulation, in which actions are inferred from forecast visual states \cite{tian2024seer}. CoT-VLA \cite{zhao2025cotvla} predicts future visual goals as an explicit visual chain of thought before producing actions. WorldVLA unifies image and action generation in one autoregressive model \cite{cen2025worldvla}. V-JEPA 2-AC \cite{assran2025vjepa2} instead predicts future states in a learned representation space and uses them for planning. Genie Envisioner builds policy learning and neural simulation on a shared video diffusion foundation \cite{liao2025genieenvisioner}.
DreamZero \cite{ye2026dreamzero} builds a large WAM on a pretrained video diffusion backbone and demonstrates strong physical generalization. $\tau_0$-WM  \cite{zhou2026tau0wm} combines video-action prediction, action-conditioned simulation, and candidate evaluation in one framework. GigaWorld-Policy \cite{ye2026gigaworldpolicy} makes future-video generation optional at deployment and places the action stream before the video stream. 
Flash-WAM \cite{akbari2026flashwam} distills iterative video-action diffusion into very few sampling steps. Efficient-WAM \cite{li2026efficientwam} uses a compact video expert, sparse video tokens, and asymmetric denoising. AGRA \cite{qiu2026agra} shows that visually plausible futures do not always yield accurate actions and aligns video features with action-relevant regions.

Video-generation-based WAMs couple robot control to a substantially harder surrogate problem. A robot ultimately executes a low-dimensional action sequence, but the model must first process or synthesize dense multi-frame latents. Most intermediate frames describe visual transition. They do not directly specify the final physical state that determines the next action. This mismatch increases sequence length, memory use, and training cost. During inference, iterative denoising over future video and action latents introduces latency into the control loop. More importantly, features optimized for visual reconstruction are not necessarily organized for low-level control, as recent representation analyses have shown \cite{qiu2026agra}. Distillation, token sparsification, and smaller video backbones reduce this cost, but they retain video synthesis as the underlying objective. $\mathtt{DELE}$-$\mathtt{w0.5}$ instead formulates WAM as future-state and action prediction. It predicts a compact future state that is causally relevant to manipulation and infers the corresponding action from this state. The model does not rely on any video generation technique and does not reconstruct intermediate frames. It therefore avoids the high-dimensional visual redundancy of video WAMs and provides a more direct, cheaper, and lower-latency route from world prediction to robot control.

\section{Preliminary}
\subsection{Robotic Manipulation}

We model robotic manipulation as a probabilistic inference model.
For each time $t$, the robot obtains the multi-view RGB images $\bO_t$ (also known as observation), the proprioception $\bq_t$, and language instruction $\mathbf{L}$, we aim to predict the action according to the policy
$
\bA_{t}\sim\pi(\cdot|\bO_t,\bq_t,\mathbf{L}),
$
where 
\[
\bA_{t}=[\ba_{t+1},\ba_{t+2},\cdots,\ba_{t+H}]
\]
corresponds to an action chunk of furture actions (we use chunk  length $H=60$ for our tasks) for the robot will play;
where for the single-arm manipulation,  each $\ba\in \R^{7}$ denoted as $7$-DoF action space: a $3$-DoF relative positional displacement $(x, y, z)$, a $3$-DoF Euler angle rotation ($\mathtt{roll}$, $\mathtt{pitch}$, $\mathtt{yaw}$), and a 1-DoF binary gripper state $g \in \{0, 1\}$.
For the dual-arm manipulation, it extends the setting to a 14-DoF space.

\subsection{Language and Vision Tokenization} 
The language instruction $\bL$ defines the task to be performed by the robot (e.g. "
$\mathtt{put~the~flowers~into~the~vase}$"), where in this paper, $\bL$ is tokenized by Qwen3 \cite{qwen3technicalreport2025}, we denote it as follows,
\begin{flalign}
\label{text-encoder-01}
\hat{\bl}=\mathtt{Text~Encoder}(\bL).
\end{flalign}
At each time $t$, the current observation $\bO_{t}\in\R^{v\times3\times h\times w}$ contains multi-view images. Our robot uses three views ($v=3$): one head-mounted camera and one wrist-mounted camera on each arm.
Furthermore, a vision tokenization encoder compress the raw image $\bO_{t}$ into a latent space, denoted it as 
\begin{flalign}
\label{vision-encoder-01}
\hat{\bz}_{t}=\mathtt{Vision~Encoder}(\bO_t),
\end{flalign}
where $\hat{\bz}_{t}\in\R^{c'\times h'\times w'}$ is with the channel number of $c'$, and the latent size is with $h'\times w'$.
In this paper, we apply DINO-v3 \cite{siméoni2025dinov3} as the vision encoder to encode the input images.
The DINO-v3 encoder provides more comprehensive representations with concurrently discern spatial and geometric relationships across observations. Such representations serve as high-quality inputs for downstream policy networks (e.g., diffusion policies or action models), enhancing the accurate perception of the relative pose between the manipulator's end-effector and the target object, and consequently enabling the generation of more refined motion trajectories.

\subsection{Timestep Embedding}

This section present the method of timestep embedding. 
For a given timestep $\tau$, first, the scalar $\tau$ is embedded into a high-dimensional, multi-scale space that exhibits favorable distance properties;
then, a two-layer MLP adapts the frequency encoding for the downstream task.
Let $d$ be the embedding dimension, $k = \lfloor d/2 \rfloor$, $\tilde{\bt}\in\R^{d}$, for each $j\in\{0,1,\cdots, d-1\}$:
\begin{flalign}
\tilde{\bt}[j] =
\begin{cases} 
\cos\left(\tau \cdot 10000^{-j/k}\right), & \text{if}~0 \leq j < k, \\
\sin\left(\tau \cdot 10000^{-(j-k)/k}\right), & \text{if}~k \leq j < 2k, \\
0, & \text{if}~ j = d-1 ~\&~d \text{ is~an~odd~number}.
\end{cases}
\end{flalign}
Furthermore, we apply a two-layer MLP to map the vector $\tilde{\bt}$ from $d$-dimension to $d$-dimension as follows, 
\begin{flalign}
\mathbf{t}=\mathtt{FNN}(\tilde{\bt}),
\end{flalign}
where $\mathtt{FNN}$ is a neural network that first increases the dimensiona of $\tilde{\bt}$, and then reduces it to be the $d$-dimension.
Finally, we present above embedding as the following formulation,
\begin{flalign}
\label{timestep-embedding}
\mathbf{t}=\mathtt{Timestep~Embedding}(\tau).
\end{flalign}

\input{sections/framework.tex}

\section{Framwork of DELE-w0.5}

The framwork of $\mathtt{DELE}$-$\mathtt{w0.5}$ adopts a dual-stream attention mechanism architecture, which contains two blocks: condition stream and noise stream.

\subsection{Condition Stream Block}

\subsubsection{Input and Tokenization}

For each time $t$, we encode the language instruction input $\bL$ and current state input $\bO_t$ according to the methods \eqref{text-encoder-01} and \eqref{vision-encoder-01} correspondingly, 
and obtain the language token $\hat{\bl}$, latent space of vision $\hat{\bz}_t$.
Since the output dimensions of the vision encoder and the language encoder are not necessarily the same, 
then we apply the projection operator to ensure that feature information $\bl$ and $\bz_t$ have the same hidden dimension,
\begin{flalign}
\bl =\mathtt{Projection}(\hat{\bl}),~~\bz_t =\mathtt{Projection}(\hat{\bz}_t),
\end{flalign}
where in this paper, we set hidden dimension with the value of $1024$, the $\mathtt{Projection}(\cdot)$ operator we used is the linear neural networks.

\subsubsection{Single-Stream Attention Block}
To integrate the information of language instruction with current observations, we employ a single-stream attention block \cite{li2019visualbert} that concatenates the multimodal tokens (i.e., $\bl $ and $\bz_t$), and feeds them into a shared Transformer encoder, where the attention block enables unified, source-agnostic cross-modal interactions. Concretely, firstly, we concatenate the language and vision tokens as follows,
\begin{flalign}
\label{con-latent-space-01}
\bZ^{\mathrm{c}}_{t}=\mathtt{cat}(\bl,\bz_t)=:[\bl,\bz_t].
\end{flalign}
Then we apply the standard self-attention block \cite{transformer-nips2017} to refine the concatenated information as follows,
\begin{flalign}
\label{con-latent-space-02}
\hat{\bZ}^{\mathrm{c}}_{t}=\mathtt{Attention}(\bZ^{\mathrm{c}}_{t}),
\end{flalign}
which exploites the complementary information gain from language and vision, 
and shows the generalization robustness in embodied artificial intelligence.

\subsubsection{AdaLN and Output of Condition Stream}

Furthermore, $\mathtt{DELE}$-$\mathtt{w0.5}$ applies the AdaLN facilitates the integration of conditioning information into the feature representation, which resides the dynamic learning of intermediate representations in response to the incoming conditioning signal, and which ia a critical prerequisite for robots to attain fine-grained control.
After the single-stream attention block,  with the fused information $\hat{\bZ}^{\mathrm{c}}_{t}$, we obtain the query, keys, values as the output of condition stream as follows,
\begin{flalign}
\label{con-qkv-q}
\bQ_{\mathrm{c}}=&\mathrm{QKNorm}\left\{\bW_{q}(\mathrm{RMSNorm}(\hat{\bZ}^{\mathrm{c}}_{t})\odot \bm{\gamma}_{\mathrm{c}})\right\},\\
\label{con-qkv-k}
\bK_{\mathrm{c}}=&\mathrm{QKNorm}\left\{\bW_{k}(\mathrm{RMSNorm}(\hat{\bZ}^{\mathrm{c}}_{t})\odot \bm{\gamma}_{\mathrm{c}})\right\},\\
\label{con-qkv-v}
\bV_{\mathrm{c}}=&\bW_{v}(\mathrm{RMSNorm}(\hat{\bZ}^{\mathrm{c}}_{t})\odot \bm{\gamma}_{\mathrm{c}}),
\end{flalign}
where $\mathrm{RMSNorm}(\cdot)$ is root mean square normalization \cite{RMSNorm2019}; $\bm{\gamma}_{\mathrm{c}}$ is the scaling factor for the condition stream with following linear structure,
\begin{flalign}
\bm{\gamma}_{\mathrm{c}}=1+\bW_{\mathrm{Adaln}} \mathbf{t}_{\mathrm{c}}+\mathbf{b}_{\mathrm{Adaln}},
\end{flalign}
$\bW_{\mathrm{Adaln}}$ and $\mathbf{b}_{\mathrm{Adaln}}$ are the learnable weights, $\mathbf{t}_{\mathrm{c}}$ is the timestep $\tau=1$ for  condition stream, i.e., 
\[\mathbf{t}_{\mathrm{c}}
\overset{\eqref{timestep-embedding}}=\mathtt{Timestep~Embedding}(1);
\]
$\mathrm{QKNorm}(\cdot)$ is a normalization operator maps any vector $\mathbf{x}$ as follows,
\begin{flalign}
\mathbf{x}'=\mathrm{QKNorm}(\mathbf{x})={\mathbf{x}}/{\|\mathbf{x}\|_2};
\end{flalign}
and $\mathrm{QKNorm}(\cdot)$ applies $\ell_2$ normalization along the head dimension of each query and key matrix prior to multiplying them.

\subsection{Noise Stream Block}

For each time $t$,  we need to add noise to the clear data $\bA_{t}$ and next state $\bO_{t+\Delta t}$.
Let $\tau\sim\mathcal{U}_{[0,1]}$, $\bepsilon\sim\calN(\bm{0},\bI)$, where $\mathcal{U}_{[0,1]}$ is the uniform distribution on $[0,1]$, 
we obtain noise action as follows,
\begin{flalign}
\bA^{\tau}_{t}=\tau \bA_{t}+(1-\tau)\bepsilon.
\end{flalign}
The noise addition of image is on the latent space:
\begin{flalign}
\label{vision-encoder-next-image-01}
\hat{\bz}_{t+\Delta t}=&\mathtt{Vision~Encoder}(\bO_{t+\Delta t}),
\end{flalign}
then we obtain the noise image as follows,
\begin{flalign}
\bz^{\tau}_{t+\Delta t}=\tau \hat{\bz}_{t+\Delta t}+(1-\tau)\bepsilon.
\end{flalign}
Furthermore, following the same steps from \eqref{con-latent-space-01}, \eqref{con-latent-space-02}, \eqref{con-qkv-q}, \eqref{con-qkv-k} and \eqref{con-qkv-v}, we obtain the query, keys, values as the output of noise stream, denoted them as 
$\bQ_{\mathrm{n}}$, $\bK_{\mathrm{n}}$ and $\bV_{\mathrm{n}}$.

\subsection{Flow Matching Objective}

We concatenate the information from condition stream and noise stream as follows,
\begin{flalign}
\bQ=\mathtt{cat}(\bQ_{\mathrm{c}},\bQ_{\mathrm{n}}),~~~\bK=\mathtt{cat}(\bK_{\mathrm{c}},\bK_{\mathrm{n}}),~~~\bV=\mathtt{cat}(\bV_{\mathrm{c}},\bV_{\mathrm{n}}).
\end{flalign}
 With the joint attention block and projection layer, we obtain the predicted action chunk $\hat{\bA}_{t}$, and predicted next latent  observations $\hat{\bz}_{t+\Delta t}$ as follows,
 \begin{flalign}
 \hat{\bA}^{\tau}_{t},~\hat{\bz}^{\tau}_{t+\Delta t}=\mathtt{Prejection}(\mathtt{
 Masked~Attention}(\bQ,\bK,\bV)).
 \end{flalign}
 Finally, we regress the velocity field according the following way,
 \begin{flalign}
 \mathcal{L}_{\mathrm{a}}=&\mathbb{E}_{\tau\sim\mathcal{U}_{[0,1]},\bepsilon\sim \calN(\bm{0},\bI)}\left[\left\|\mathbf{u}_{\mathrm{a}}(\bA^{\tau}_{t},\bz^{\tau}_{t+\Delta t},\bq_{t},\mathbf{l})-(\bA_t-\bepsilon)\right\|_{2}^{2}
\right],
\\
 \mathcal{L}_{\mathrm{o}}=&\mathbb{E}_{\tau\sim\mathcal{U}_{[0,1]},\bepsilon\sim \calN(\bm{0},\bI)}\left[\left\|\mathbf{o}_{\mathrm{a}}(\bA^{\tau}_{t},\bz^{\tau}_{t+\Delta t},\bq_{t},\mathbf{l})-(\hat{\bz}_{t+\Delta t}-\bepsilon)\right\|_{2}^{2}
\right].
 \end{flalign} 
 The overall training objective is 
  \begin{flalign}
   \mathcal{L}=
 \mathcal{L}_{\mathrm{a}}+ \lambda\mathcal{L}_{\mathrm{o}},
 \end{flalign} 
 where $\lambda$ balances the action learning the next state prediction.
%
%

\newcommand{\maskFigureFont}{\small}

\newcommand{\maskVisibleCell}[4]{%
  \draw[rounded corners=1.8pt, line width=0.8pt, draw=#4, fill=#3]
    ({#1+0.1008},{#2+0.1008}) rectangle ++(0.5184,0.5184);%
}

\newcommand{\maskEmptyCell}[2]{%
  \draw[rounded corners=1.8pt, line width=0.8pt, draw=gray, fill=white]
    ({#1+0.1008},{#2+0.1008}) rectangle ++(0.5184,0.5184);%
}

\newcommand{\maskTextLabel}[5]{%
  \node[anchor=#5, font=\maskFigureFont, text=#4, align=center]
    at (#1,#2) {#3};%
}

\newcommand{\trainingMaskPanel}[2]{%
\begin{scope}[shift={(#1,#2)}]
  \node[font=\maskFigureFont] at (1.12128,2.8) {Training};

  \maskTextLabel{0.3592}{2.54}{$\mathbf{L}$}{red}{center}
  \maskTextLabel{0.93392}{2.54}{$\mathbf{O}_{t}$}{blue}{center}
  \maskTextLabel{1.50864}{2.54}{$\mathbf{A}_{t}$}{brown}{center}
  \maskTextLabel{2.08336}{2.54}{$\mathbf{O}_{t+\Delta t}$}{blue}{center}

  \maskTextLabel{-0.12}{1.98336}{$\mathbf{L}$}{red}{east}
  \maskTextLabel{-0.12}{1.40864}{$\mathbf{O}_{t}$}{blue}{east}
  \maskTextLabel{-0.12}{0.83392}{$\mathbf{A}_{t}$}{brown}{east}
  \maskTextLabel{-0.12}{0.2592}{$\mathbf{O}_{t+\Delta t}$}{blue}{east}

  \maskVisibleCell{0.00}{1.72416}{red!38}{red}
  \maskVisibleCell{0.57472}{1.72416}{red!38}{red}
  \maskEmptyCell{1.14944}{1.72416}
  \maskEmptyCell{1.72416}{1.72416}

  \maskVisibleCell{0.00}{1.14944}{blue!38}{blue}
  \maskVisibleCell{0.57472}{1.14944}{blue!38}{blue}
  \maskEmptyCell{1.14944}{1.14944}
  \maskEmptyCell{1.72416}{1.14944}

  \maskVisibleCell{0.00}{0.57472}{brown!38}{brown}
  \maskVisibleCell{0.57472}{0.57472}{brown!38}{brown}
  \maskVisibleCell{1.14944}{0.57472}{brown!38}{brown}
  \maskEmptyCell{1.72416}{0.57472}

  \maskVisibleCell{0.00}{0.00}{blue!38}{blue}
  \maskVisibleCell{0.57472}{0.00}{blue!38}{blue}
  \maskVisibleCell{1.14944}{0.00}{blue!38}{blue}
  \maskVisibleCell{1.72416}{0.00}{blue!38}{blue}
\end{scope}%
}

\newcommand{\inferenceMaskPanel}[2]{%
\begin{scope}[shift={(#1,#2)}]
  \node[font=] at (0.83392,2.8) {Inference};

  \maskTextLabel{0.3592}{2.54}{$\mathbf{L}$}{red}{center}
  \maskTextLabel{0.93392}{2.54}{$\mathbf{O}_{t}$}{blue}{center}
  \maskTextLabel{1.50864}{2.54}{$\mathbf{A}_{t}$}{brown}{center}

  \maskTextLabel{-0.12}{1.98336}{$\mathbf{L}$}{red}{east}
  \maskTextLabel{-0.12}{1.40864}{$\mathbf{O}_{t}$}{blue}{east}
  \maskTextLabel{-0.12}{0.83392}{$\mathbf{A}_{t}$}{brown}{east}

  \maskVisibleCell{0.00}{1.72416}{red!38}{red}
  \maskVisibleCell{0.57472}{1.72416}{red!38}{red}
  \maskEmptyCell{1.14944}{1.72416}

  \maskVisibleCell{0.00}{1.14944}{blue!38}{blue}
  \maskVisibleCell{0.57472}{1.14944}{blue!38}{blue}
  \maskEmptyCell{1.14944}{1.14944}

  \maskVisibleCell{0.00}{0.57472}{brown!38}{brown}
  \maskVisibleCell{0.57472}{0.57472}{brown!38}{brown}
  \maskVisibleCell{1.14944}{0.57472}{brown!38}{brown}
\end{scope}%
}

\begin{figure}[t]
  \centering
  \begin{tikzpicture}[x=2.5cm,y=2.5cm]
    \trainingMaskPanel{0}{0}
    \inferenceMaskPanel{3.30}{0}
  \end{tikzpicture}
  \caption{\maskFigureFont\textbf{Attention masks for training and inference.}
  Rows are query-token groups and columns are key/value-token groups.
  $\mathbf{L}$, $\mathbf{O}_{t}$, $\mathbf{A}_{t}$, and $\mathbf{O}_{t+\Delta t}$ denote language, current observation, action, and predicted future observation, respectively.
  During training (left), $\mathbf{L}$ and $\mathbf{O}_{t}$ attend bidirectionally; $\mathbf{A}_{t}$ attends to $\{\mathbf{L},\mathbf{O}_{t},\mathbf{A}_{t}\}$; and $\mathbf{O}_{t+\Delta t}$ attends to all groups.
  During inference (right), $\mathbf{O}_{t+\Delta t}$ is omitted.
  Colored and white cells indicate permitted and masked attention, respectively.
  Multi-view tokens are merged into each observation group, and padded language positions are masked along both dimensions.}
  \label{fig:attention-mask}
\end{figure}
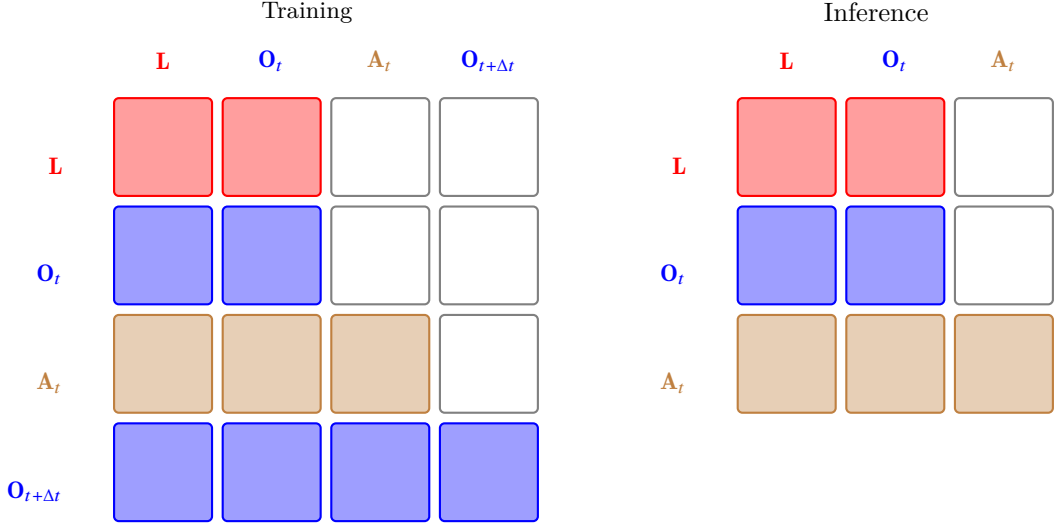

 \subsection{Training and Inference}
 
During training, the input tokens are grouped as language, current multi-view observation, action, and future observation.
The language and current-observation tokens form the conditioning group and attend bidirectionally to one another.
They cannot attend to either the action tokens or the predicted future-observation tokens, which prevents information from the prediction targets from leaking into the conditioning representation.
The action tokens attend to the entire conditioning group and to the action group itself, but they cannot access the predicted future observations.
Therefore, robot actions are learned only from the language instruction, the current multi-view observations, and the dependencies within the action sequence.
In contrast, the predicted future-observation tokens attend to all token groups, allowing the model to learn the physical state transition conditioned on the instruction, current observations, and robot actions.
During inference, the two predicted future-observation groups are removed, while the visibility rules for the conditioning and action groups remain unchanged.
The model therefore predicts the action sequence without generating any future visual tokens, which shortens the inference sequence, removes unnecessary visual generation, and enables efficient low-latency robot control.

\section{Experiments}
\label{sec:experiments}

In this section, we study the following three questions: How reliably does
$\mathtt{DELE}$-$\mathtt{w0.5}$ complete real-world manipulation tasks with different horizons? How
does it compare with representative vision-language-action (VLA) policies
under the same fine-tuning data and evaluation protocol? At which stages do
the methods fail as task horizon and coordination difficulty increase? We
answer these questions through full-task success, normalized stage progress,
and task-wise stage-reach analysis.

\subsection{Experimental Setup}
\label{sec:experimental_setup}

\begin{figure*}[t]
    \centering
    \includegraphics[width=\textwidth]{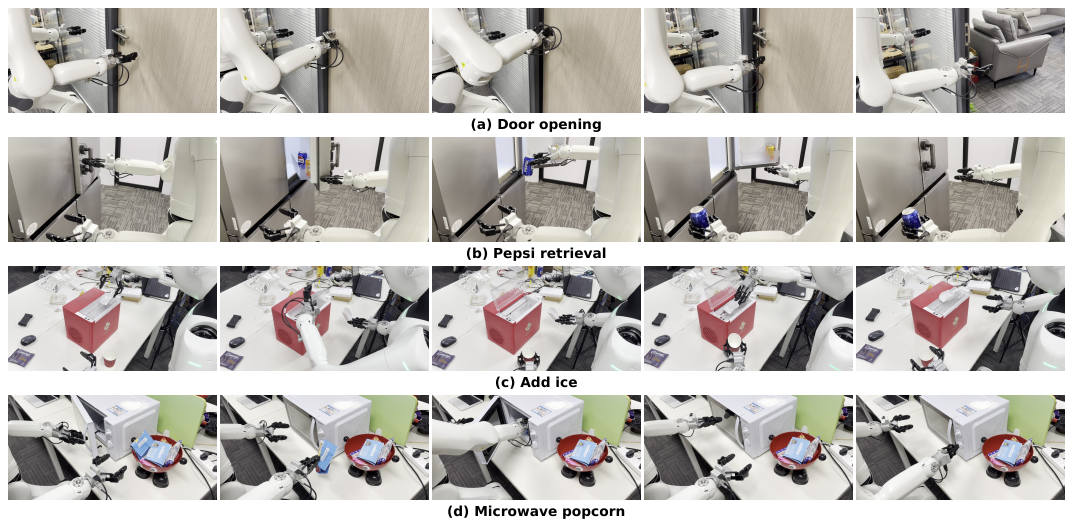}
    \caption{Representative stage sequences for the four real-robot tasks. Each
    row shows five temporally ordered key states from left to right for door
    opening, Pepsi retrieval, adding ice, and microwave popcorn.}
    \label{fig:task_scenes}
\end{figure*}

\paragraph{Robot and tasks.}
We conduct all experiments on the same Astribot S1 dual-arm robot in fixed
real-world scenes. We consider four tasks with different manipulation horizons:
opening a hinged door, retrieving a canned Pepsi from a refrigerator, adding
ice to a cup, and loading a popcorn package into a microwave and starting the
heating cycle. Figure~\ref{fig:task_scenes} illustrates the task sequences, and
Table~\ref{tab:task_definitions} gives the ordered stages and conditions for
full success. A stage is counted only when all preceding stages have been
completed in order. Thus, a later action cannot compensate for a skipped
prerequisite.

\paragraph{Task-suite design.}
The task suite covers six sources of difficulty in long-horizon manipulation:
contact-rich articulated interaction, target-object retrieval, bimanual
transfer, tool use, constrained insertion, and terminal-state verification.
The door task requires a stable handle grasp followed by timely handle rotation
and pushing to release the latch. The Pepsi task adds target extraction, a
right-to-left handoff, and refrigerator closure while retaining the can. The
ice task is the longest sequence and requires coordinated use of the scoop,
ice-maker lid, and cup, followed by pouring and scene restoration. The
microwave task combines appliance opening, package pickup, constrained
insertion, door closure, and heating activation. These tasks therefore test
not only individual manipulation skills, but also whether a policy can connect
the skills while preserving previously achieved task state.

\begin{table*}[t]
    \centering
    \caption{Ordered task stages and complete task conditions. Stage zero
    denotes no effective progress, and reaching the final stage denotes full
    task completion.}
    \label{tab:task_definitions}
    \begingroup
    \small
    \setlength{\tabcolsep}{5pt}
    \renewcommand{\arraystretch}{1.12}
    \begin{tabularx}{\textwidth}{@{}
        >{\raggedright\arraybackslash}p{0.17\textwidth}
        >{\centering\arraybackslash}p{0.035\textwidth}
        >{\raggedright\arraybackslash}X
        >{\raggedright\arraybackslash}p{0.235\textwidth}@{}}
        \toprule
        \textbf{Task} & \textbf{$K$} & \textbf{Ordered stages} &
        \textbf{Complete physical state} \\
        \midrule
        \textbf{Door opening} & 4
            & Handle contact $\rightarrow$ stable grasp $\rightarrow$ coordinated
              handle rotation and push $\rightarrow$ sustained opening
            & Door remains beyond $45^\circ$ for at least 3 seconds. \\
        \addlinespace[2pt]
        \textbf{Pepsi retrieval} & 5
            & Handle interaction $\rightarrow$ door open $\rightarrow$ can extracted
              $\rightarrow$ left-hand handoff $\rightarrow$ door closed
            & Left gripper retains the target can while the refrigerator is closed. \\
        \addlinespace[2pt]
        \textbf{Add ice} & 6
            & Scoop pickup $\rightarrow$ lid open $\rightarrow$ cup pickup
              $\rightarrow$ ice acquired $\rightarrow$ ice poured $\rightarrow$
              scene restored
            & Ice remains in the upright cup and the manipulated scene is restored. \\
        \addlinespace[2pt]
        \textbf{Microwave popcorn} & 5
            & Door open $\rightarrow$ package pickup $\rightarrow$ insertion
              $\rightarrow$ door closed $\rightarrow$ heating started
            & Popcorn is inside the closed microwave and heating visibly starts. \\
        \bottomrule
    \end{tabularx}
    \endgroup
\end{table*}

\paragraph{Baselines and training protocol.}
We compare $\mathtt{DELE}$-$\mathtt{w0.5}$ with seven representative VLA policies:
GigaWorld-Policy-0.5 (GWP0.5)~\cite{gigaworld2026gwp05},
Xiaomi Robotics-0 (XR0)~\cite{cai2026xiaomirobotics0},
$\pi_{0.5}$~\cite{intelligence2025pi05},
LingBot-VLA2~\cite{wu2026vla2}, Hy-Embodied-0.5-VLA
(HY-VLA)~\cite{zhang2026hyembodied05vla}, Spirit-v1.5~\cite{spiritai2026spiritv15}, and
GR00T-N1.7~\cite{nvidia2025gr00tn1}. All methods are compared under identical
fine-tuning data and evaluation protocols. Each baseline follows its official
training configuration and is trained on the same task dataset for an
equivalent of three epochs. At evaluation time, all methods use the same robot
embodiment, task instructions, scene-reset configurations, 180-second
execution budget, ordered-stage definitions, and complete-success criteria.

\paragraph{Evaluation protocol.}
For each method-task pair, we run twenty formal trials, giving
$8 \times 4 \times 20 = 640$ real-robot trials in total. Before each trial, we
reset the robot and all task-relevant objects to predefined poses. Each policy
has at most 180 seconds to complete the task. The operator provides no physical
assistance after execution begins and stops a rollout only after observable
success or terminal task failure, or when continued execution is unsafe. A
rollout terminated for unsafe execution is counted as a failure. A trial is
counted as a full success only when the task-specific complete condition in
Table~\ref{tab:task_definitions} is satisfied without human assistance.

\paragraph{Metrics.}
We report two metrics. The first is the \emph{full-task success rate}, which
measures whether the complete physical task condition is satisfied. The second
is normalized ordered-stage progress. For a task with $K$ stages and a trial
whose highest consecutively completed stage is $s$, the progress score is
$s/K$. We average this score over twenty trials for each task and then
macro-average over the four tasks. Therefore, every task has equal weight even
though the tasks contain different numbers of stages. For Add Ice, the eight
raw annotation stages are consolidated into six semantic stages in the paper:
raw stages 5--7 map to semantic stage 5, and raw stage 8 maps to semantic
stage 6. The raw trial annotations remain unchanged.

\subsection{Main Results}
\label{sec:main_results}

\begin{figure*}[t]
    \centering
    \includegraphics[width=\textwidth]{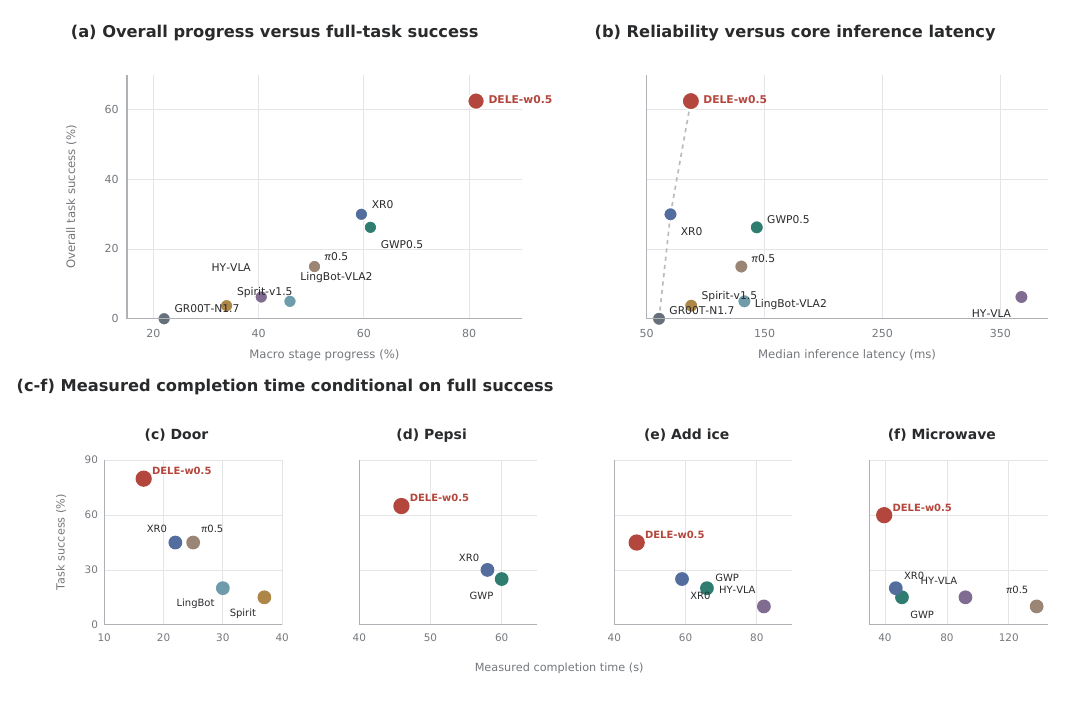}
    \caption{Performance and efficiency overview. (a) Macro-average ordered-stage
    progress versus overall full-task success. (b) Overall success versus median
    core-model inference latency on an NVIDIA RTX 4090. (c--f) Task-level
    success versus measured completion-time centers conditional on successful
    completion. Method abbreviations in the compact task panels
    correspond to the full names in Table~\ref{tab:main_real_results} and panels
    (a)--(b).}
    \label{fig:main_results}
\end{figure*}

Table~\ref{tab:main_real_results} and Figure~\ref{fig:main_results} compare
normalized progress and full-task success. Results show that $\mathtt{DELE}$-$\mathtt{w0.5}$ performs
best on all four tasks. It obtains 81.3\% macro progress and 62.5\% overall
success (50/80). GWP0.5 is the strongest baseline in macro progress at 61.3\%,
whereas XR0 has the highest baseline full-task success at 30.0\% (24/80).
Using unrounded values and the strongest baseline for each metric, $\mathtt{DELE}$-$\mathtt{w0.5}$
improves macro progress by 20.1 percentage points and full-task success by
32.5 percentage points.

\begin{table*}[t]
    \centering
    \caption{Real-robot results over twenty trials per task. Each task entry is
    reported as normalized progress (Prog.) and full-task success (Succ.), in
    percent. Macro progress weights the four tasks equally; overall success is
    computed over all 80 trials per method.}
    \label{tab:main_real_results}
    \begingroup
    \footnotesize
    \setlength{\tabcolsep}{2.6pt}
    \renewcommand{\arraystretch}{1.12}
    \sisetup{
        mode=text,
        reset-text-series=false,
        text-series-to-math=true,
        table-format=2.1,
        table-number-alignment=center
    }
    \begin{tabular*}{\textwidth}{@{\extracolsep{\fill}}l*{10}{S}@{}}
        \toprule
        & \multicolumn{2}{c}{\textbf{Door}}
        & \multicolumn{2}{c}{\textbf{Pepsi}}
        & \multicolumn{2}{c}{\textbf{Add ice}}
        & \multicolumn{2}{c}{\textbf{Microwave}}
        & \multicolumn{1}{c}{\textbf{Macro}}
        & \multicolumn{1}{c}{\textbf{Overall}} \\
        \cmidrule(lr){2-3}\cmidrule(lr){4-5}\cmidrule(lr){6-7}
        \cmidrule(lr){8-9}\cmidrule(lr){10-10}\cmidrule(l){11-11}
        \textbf{Method}
        & \multicolumn{1}{c}{Prog.} & \multicolumn{1}{c}{Succ.}
        & \multicolumn{1}{c}{Prog.} & \multicolumn{1}{c}{Succ.}
        & \multicolumn{1}{c}{Prog.} & \multicolumn{1}{c}{Succ.}
        & \multicolumn{1}{c}{Prog.} & \multicolumn{1}{c}{Succ.}
        & \multicolumn{1}{c}{Prog.} & \multicolumn{1}{c}{Succ.} \\
        \midrule
        \textbf{$\mathtt{DELE}$-$\mathtt{w0.5}$ (ours)}
            & \bfseries 95.0 & \bfseries 80.0
            & \bfseries 82.0 & \bfseries 65.0
            & \bfseries 63.3 & \bfseries 45.0
            & \bfseries 85.0 & \bfseries 60.0
            & \bfseries 81.3 & \bfseries 62.5 \\
        \addlinespace[1pt]
        GWP0.5
            & 80.0 & 45.0
            & 60.0 & 25.0
            & 50.0 & 20.0
            & 55.0 & 15.0
            & 61.3 & 26.3 \\
        XR0
            & 77.5 & 45.0
            & 67.0 & 30.0
            & 41.7 & 25.0
            & 52.0 & 20.0
            & 59.5 & 30.0 \\
        $\pi_{0.5}$
            & 80.0 & 45.0
            & 42.0 & 5.0
            & 32.5 & 0.0
            & 48.0 & 10.0
            & 50.6 & 15.0 \\
        LingBot-VLA2
            & 60.0 & 20.0
            & 50.0 & 0.0
            & 25.8 & 0.0
            & 48.0 & 0.0
            & 46.0 & 5.0 \\
        HY-VLA
            & 57.5 & 0.0
            & 35.0 & 0.0
            & 37.5 & 10.0
            & 32.0 & 15.0
            & 40.5 & 6.3 \\
        Spirit-v1.5
            & 57.5 & 15.0
            & 35.0 & 0.0
            & 14.2 & 0.0
            & 29.0 & 0.0
            & 33.9 & 3.8 \\
        GR00T-N1.7
            & 40.0 & 0.0
            & 25.0 & 0.0
            & 9.2 & 0.0
            & 14.0 & 0.0
            & 22.0 & 0.0 \\
        \bottomrule
    \end{tabular*}
    \endgroup
\end{table*}

Figure~\ref{fig:main_results}(a) shows that $\mathtt{DELE}$-$\mathtt{w0.5}$ not only reaches farther
through the ordered task stages, but also converts intermediate progress into
complete execution more consistently. Several baselines reach early
interactions but lose progress during object transfer, bimanual coordination,
or final-state completion. Figure~\ref{fig:main_results}(b) further shows that
$\mathtt{DELE}$-$\mathtt{w0.5}$ achieves 62.5\% overall success with a median inference time of
\SI{87.5}{ms}. It is faster per inference call than GWP0.5, $\pi_{0.5}$,
LingBot-VLA2, and HY-VLA, and is comparable to Spirit-v1.5 while achieving
substantially higher success. Core-model latency excludes network transfer,
image decoding, action projection, IK, trajectory processing, and robot
execution.

Figures~\ref{fig:main_results}(c)--(f) compare task success with measured
completion time conditional on successful completion. Across its 16, 13, 9,
and 12 successful
trials, $\mathtt{DELE}$-$\mathtt{w0.5}$ has measured mean completion times of \SI{16.66}{s},
\SI{45.92}{s}, \SI{46.27}{s}, and \SI{39.50}{s} for Door, Pepsi, Add Ice,
and Microwave, respectively. The same runtime measurement protocol is used for
all methods. Thus, the comparison preserves the distinction between
model computation, successful robot execution, and success probability.

\subsection{Task-wise and Stage-wise Analysis}
\label{sec:stage_analysis}

\begin{figure*}[t]
    \centering
    \includegraphics[width=\textwidth]{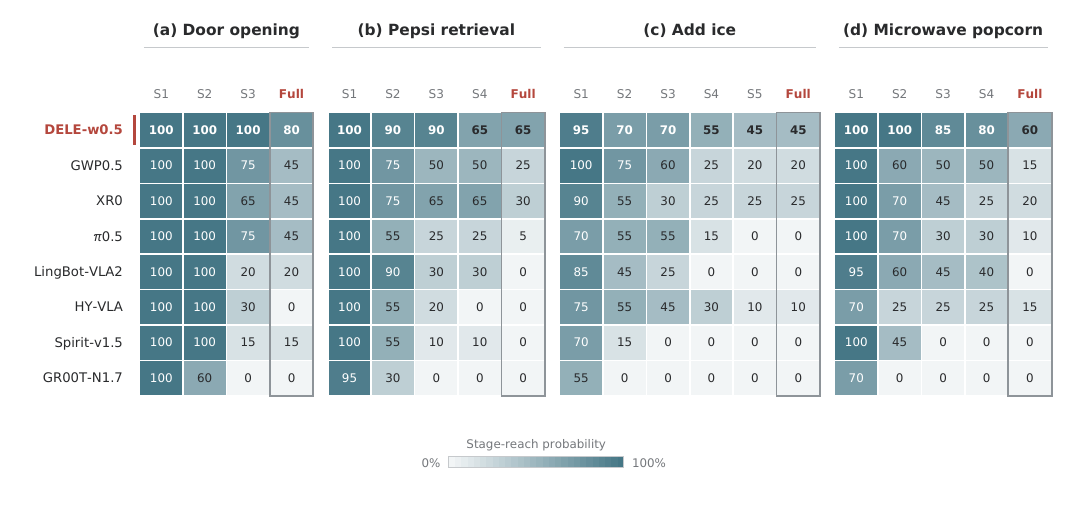}
    \caption{Stage-wise long-horizon analysis. Each cell reports the percentage
    of twenty trials whose ordered-prefix score reaches or exceeds the indicated
    stage. Task blocks share a common model axis and probability scale;
    \emph{Full} represents complete physical task success.}
    \label{fig:stage_reach}
\end{figure*}

Figure~\ref{fig:stage_reach} shows the empirical stage-reach probability
$P(s \geq k)$ for each task. A single progress value measures how far a method
moves on average, while the stage-retention maps show the exact stage at which performance
drops. We make four observations from the results.

\paragraph{Door opening.}
From Figure~\ref{fig:stage_reach}(a), $\mathtt{DELE}$-$\mathtt{w0.5}$ achieves 16/20 full successes
and 95.0\% normalized progress. GWP0.5 and $\pi_{0.5}$ each obtain 80.0\%
progress, while GWP0.5, XR0, and $\pi_{0.5}$ each complete 9/20 trials. After
grasping the handle, the robot must maintain the grasp,
rotate the handle to a sufficient angle, and push at the correct time. $\mathtt{DELE}$-$\mathtt{w0.5}$
reaches this coordinated rotation-and-push stage in all twenty trials, and its
four failures occur only at the final sustained-opening condition. In
comparison, most baselines reach the handle but fail to synchronize the
rotation and push or fail to maintain the required opening angle. This result
shows that handle perception alone is insufficient; the task depends on
timely contact-rich coordination.

\paragraph{Pepsi retrieval.}
Figure~\ref{fig:stage_reach}(b) shows that $\mathtt{DELE}$-$\mathtt{w0.5}$ completes 13/20 Pepsi
retrieval trials. It opens the refrigerator and extracts the target can in
18/20 trials. The remaining failures mainly occur at the right-to-left handoff
and final door closure. XR0 is the strongest baseline on this task, with 67.0\%
progress and 6/20 full successes; GWP0.5 follows with 60.0\% progress and 5/20
successes. LingBot-VLA2 reaches a stable left-hand handoff in six trials but
never closes the refrigerator while retaining the can, whereas $\pi_{0.5}$
completes the full task once. Thus, articulated interaction is an early
bottleneck, while handoff and closure determine whether intermediate progress
becomes a complete retrieval.

\paragraph{Adding ice.}
Figure~\ref{fig:stage_reach}(c) reports the longest task sequence. $\mathtt{DELE}$-$\mathtt{w0.5}$
completes the six-stage task in 9/20 trials. It picks up the scoop in 19 trials,
reaches cup pickup in 14, acquires ice in 11, and pours ice into the cup in
nine. GWP0.5 is the strongest baseline by progress (50.0\%), while XR0 has the
highest baseline success count (5/20). HY-VLA reaches 37.5\% progress and
completes the task twice. $\pi_{0.5}$ obtains 32.5\% progress but never places
ice into the cup. These results show that a successful initial grasp is not
enough: repeated bimanual coordination, tool use, precise release, and scene
restoration determine final performance.

\paragraph{Microwave popcorn.}
From Figure~\ref{fig:stage_reach}(d), $\mathtt{DELE}$-$\mathtt{w0.5}$ obtains 12/20 full successes and
85.0\% normalized progress. It opens the microwave and picks up the package in
every trial, closes the microwave in 16 trials, and starts heating in 12.
GWP0.5 and XR0 reach 55.0\% and 52.0\% progress, with 3/20 and 4/20 full
successes. Although $\pi_{0.5}$ and Spirit-v1.5 open the microwave in all
twenty trials and LingBot-VLA2 does so in 19, their final success rates are only
2/20, 0/20, and 0/20. This stage-wise drop explains why appliance opening alone
is a weak measure of long-horizon performance. Package retention, insertion,
door closure, and heating activation introduce successive opportunities for
failure.

\subsection{Failure Analysis}
\label{sec:failure_analysis}

For each unsuccessful trial, we record the first uncompleted ordered stage.
This view identifies the main bottleneck without relying on model-specific
runtime logs. The dominant failure stage differs across tasks. For door
opening, most baselines grasp the handle but fail during coordinated rotation
and pushing. For Pepsi retrieval, weaker methods fail before full extraction,
while stronger baselines such as GWP0.5 and XR0 lose progress during handoff or
final refrigerator closure. For adding ice, the main drop occurs before ice
acquisition and pouring. For microwave popcorn, many rollouts open the
appliance but fail during insertion, door closure, or heating activation.

All four bottlenecks require a policy to preserve previously achieved task
state while changing the active arm, object, or interaction mode. $\mathtt{DELE}$-$\mathtt{w0.5}$
reaches the late stages substantially more often, although its remaining
failures still occur during object transfer, precise release, or the final
task condition. The results therefore show that long-horizon performance
depends both on individual skills and on the reliability of the transitions
between them.

\subsection{Early signs of emergence.}
In additional intervention rollouts, we observe adaptive behaviors at
intermediate states absent from the task demonstrations. When the microwave
door is closed during execution, $\mathtt{DELE}$-$\mathtt{w0.5}$ retains the popcorn package,
reopens the door, and continues toward the original task objective. When the
available opening becomes narrow, the policy instead uses the package already
in hand to push the door open, rather than returning to the demonstrated
handle-opening routine. These behaviors exhibit goal-conditioned behavioral
recomposition: the policy preserves the intended outcome while adapting its
interaction strategy to an unseen task state. We view these observations as
early evidence rather than a quantitative claim of emergence.

\section{Conclusion}

In this work, we revisit the objective of World-Action Models for robotic manipulation.
We show that video generation is not a necessary component of world modeling for robot control.
Instead of reconstructing how the world visually evolves over time, a practical world model should predict how the physical world will change after an action is executed.
Based on this observation, we introduce $\mathtt{DELE}$-$\mathtt{w0.5}$, a new future-state-driven world-action framework that infers robot actions from compact action-relevant future states without relying on video generation.
By treating future-state prediction as the intermediate representation between world understanding and action generation, $\mathtt{DELE}$-$\mathtt{w0.5}$ removes unnecessary visual redundancy and enables efficient training and low-latency inference.
Extensive real-world experiments demonstrate the effectiveness of this formulation on challenging long-horizon manipulation tasks.
Across 640 robot trials on four diverse tasks, $\mathtt{DELE}$-$\mathtt{w0.5}$ consistently outperforms representative VLA baselines, achieving 62.5\% full-task success and 81.3\% normalized ordered-stage progress.
The results show that the advantage of $\mathtt{DELE}$-$\mathtt{w0.5}$ becomes more significant as task horizons increase and errors accumulate across multiple interaction stages.
We further observe early signs of emergence under unseen intermediate-state interventions, where the policy preserves the task objective and composes an alternative interaction strategy.
These findings suggest that action-relevant future-state prediction provides a more direct and practical foundation for embodied world models, offering a promising direction toward scalable and deployable robot intelligence.

\bibliographystyle{plainnat}
\bibliography{references}

\end{document}